\documentclass{article}
\usepackage{arxiv}
\pdfoutput=1
\usepackage[utf8]{inputenc} 
\usepackage[T1]{fontenc}    
\usepackage{hyperref}       
\usepackage{url}            
\usepackage{booktabs}       
\usepackage{amsfonts}       
\usepackage{nicefrac}       
\usepackage{microtype}      
\usepackage{amsmath}
\usepackage{algorithm}
\usepackage{algorithmic}
\usepackage{graphicx}
\usepackage{wrapfig}
\usepackage{cite}

\title{Playing Atari Ball Games with Hierarchical Reinforcement Learning}
\author{%
  Hua Huang\\
  Department of Mathematics\\
  Florida State University\\
  Tallahassee, FL 32304 \\
  \texttt{hhuang@math.fsu.edu} \\
  \And
 Adrian Barbu\thanks{Corresponding author: abarbu@stat.fsu.edu} \\
  Department of Statistics \\
  Florida State University\\
  Tallahassee, FL 32304 \\
  \texttt{abarbu@stat.fsu.edu} \\
}

\begin{document}

\maketitle

\begin{abstract}
	Human beings are particularly good at reasoning and inference from
    just a few examples. When facing new tasks, humans will leverage 
    knowledge and skills learned before, and quickly integrate them with
    the new task. In addition to learning by experimentation, human also learn 
    socio-culturally through instructions and learning by example.
    In this way humans can learn much faster compared with most
    current artificial intelligence algorithms in many tasks. In this paper,
    we test the idea of speeding up machine learning through social
    learning. We argue that in solving real-world problems, especially when 
    the task is designed by humans, and/or for humans, there are typically 
    instructions from user manuals and/or human experts which give guidelines
    on how to better accomplish the tasks. We argue that these instructions 
    have tremendous value in designing a reinforcement learning 
    system which can learn in human fashion, and we test the idea by 
    playing the Atari games Tennis and Pong. We experimentally demonstrate that
    the instructions provide key information about the task, which can be
    used to decompose the learning task into sub-systems and construct options
    for the temporally extended planning, and dramatically accelerate the 
    learning process.
\end{abstract}

\section{Introduction}
Humans are especially capable in solving new tasks by leveraging prior knowledge 
acquired by years-long interaction with the real-world, receiving systematic 
education, and learning from others. Humans are also particularly good at 
building rich internal models, which in practice are approximate but 
sufficiently accurate to guide inference and prediction\cite{lake2017building}.
When humans face a new task, like learning to drive a car 
or playing a new video game, they do not start the new task 
from scratch. For example, when one learns how to drive, first of all one needs
to go a thorough traffic rule learning phase and in most cases take a traffic rule test, 
before any driving is done. To prepare 
for the test, one needs to read the drivers' license handbook, in which there
are introductions about traffic controls such as pavement markings and traffic
control signals. Once one passes the traffic rule test and obtains the learner's permit, 
he/she is equipped with the theoretical knowledge for driving and can start to build and
sharpen his/her skills by driving a car on road. 
Furthermore, in the actual driving, one does not learn to drive from scratch, but instead
uses the knowledge and skills built while growing up to assist the task. 
For example, when learning to drive, most of us already know
Newtonian physics, basic mathematics and how to understand the world around us using our
vision. Based on the acquired skill set, we 
build a generative model for the entities involved in driving. For example,
we have a qualitative understanding of the dynamics of the car, and we can
guess the intentions of other drivers and pedestrians to act correspondingly.
These rich models dramatically accelerate the learning process. Moreover, it is typically
required by law that a new driver carrying a learner's
permit be accompanied by a licensed driver in the front passenger seat
while he or she is driving on the road. Thus most drivers learn to drive with an experienced
driver providing guidance, and some even go to driving school, where 
professional teachers teach students how to drive. When taught by a human, hundreds
of concrete and/or abstract concepts are injected into the student, which can be then leveraged
to quickly grasp the key structures of driving. By actually 
driving a car on the road, students reinforce their understanding of these
instructions. First-hand driving experience gradually gives them information on the
response of the car when different actions are taken in diverse driving
scenarios. With more practice, the learner becomes more and more experienced and is 
finally able to drive smoothly. Typically after less than 20 hours of driving,
one is ready to go anywhere in any car under any weather conditions.

Another example of learning with guidance is playing video 
games. Most of new players will first of all take a look at the user manual.
This manual will usually contain a basic introduction to the game 
environment, the rules of playing the games, the hints on how to play it well,
and even more, there might be a specially designed game mode in which 
the game speed is lower, so the player can learn how to manipulate their 
agent in an easier and more friendly game environment. Once the player got a
feeling of the game and built the basic necessary skills, they can 
switch to the normal game environment and further sharpen their game
playing skills. It's also quite common that humans will begin learning by watching videos of
professional games playing played by experts. By mimicking the expert
behavior, the gamer can quickly pickup the tactics of playing the game. 

Based on these observations, we argue that in real world tasks, which
are either designed by humans, and/or for humans, very likely there
exist "user manual" -like materials to begin with, 
and experts to learn from. There types of guidance typically provide essential 
information about the environment that humans are going to interact with,
the rules to follow, and what's more important, the guidance 
and hints on how to accomplish the designated task. Based on these hints, we can
constrain our choices and build a roadmap to accomplish the task, which 
reduces the search space dramatically, especially when the dimension of 
environment is high and the task involves many decision making steps.
In this work, we will focus on the guidance and/or hints, and apply
them within the framework of hierarchical reinforcement learning. 
Hierarchical reinforcement learning may be one of the most promising 
approaches if we want to use reinforcement learning to solve real-world 
applications, which typically have sparse rewards and long decision
processes. 

\section{Related Work}
Hierarchical reinforcement learning has been long considered appealing
for solving problems that have long decision processes in complex environments.
The learning system has to learn to represent spatial-temporal abstractions and 
explore environment efficiently. The traditional $\epsilon$-greedy policy works well
for local exploration but behaves poorly when the optimal policy involves extended
state-action sequences. Typically the goal in hierarchical reinforcement 
learning is to decompose the task into subtasks that
are easier to solve. Another consideration in taking a hierarchical approach
is that the world operates in a combinatorial way, and even better, in many 
cases we can find the entities involved obey a relatively simple law. We can
adopt a divide-and-conquer approach and use an appropriate approach for learning the subtask, 
which could be an analytical approach, a supervised learning or even an unsupervised
learning approach, model-based or model-free. Each approach has 
its strengths and weaknesses, and we can choose the appropriate one to
tackle the sub-system.

Most of the current efforts in hierarchical reinforcement learning are
focused on automatically learning the options, which are the higher level actions.
Bacon at al. \cite{bacon2017option} proposed to learn options end-to-end
automatically. Since the optimal policy is carried out by primitive actions, 
 we theoretically do not need temporally-extended options. Correspondingly,
their policies converged to either one single long option or trivial 
options that last only one step. They had to add regularization to the
reward signal to encourage temporally extended 
options. Another limitation in their work is that they assume the options apply
everywhere. The learned options can be difficult to interpret and have
no semantic meanings. Riemer et al. \cite{riemer2018learning} extended
the automatic option learning to arbitrary deep hierarchies from two
abstraction levels, and experimentally demonstrated that for significantly
complex problems, more than two levels of abstraction can be beneficial 
for learning. Machado et al. \cite{machado2017laplacian} addressed the 
option discovery problem by showing how Proto-value functions (PVFs) 
implicitly define options. PVFs capture the large-scale geometry of the 
environment, and options based on them can better explore the state space. 
When the number of states is large, and/or we cannot populate
the intermediate and/or final states, which is typical in many of the
real-world examples we are interested in, it will be challenging to
calculate the PVFs.

Roderick et al. \cite{roderick2018deep} extend the abstraction concept
to the state space. The authors argued that in many real world
situations, there are natural decompositions of the low-level state into
abstract components. Expert like human can provide the abstraction of
the state space, and the high-level abstract states can be used as goals
for lower-level agents. This state-level abstraction enhanced the
exploration for long-term planning.

Another direction related to our work is to learn a set of basic skills by pre-training, and build a
high-level policy based on these skills. Florensa et al. 
\cite{florensa2017stochastic} proposed to learn the basic
skills by pre-training, in which prior knowledge about what high level 
behaviors might be useful in the subsequent tasks are encoded in the 
rewards for the pre-training tasks. Once these skills are learned, the top-level
policy is trained to select these skills to accomplish the major task.
Riedmiller et al. \cite{riedmiller2018learning} suggested that just 
like humans have a playful phase of childhood, in which infants interact 
with environments and learn basic skills like touching and moving objects, 
and eventually build higher level skills, one can reward the agent to
explore the entities and learn basic skills when the outside reward is sparse.

There are also works that focus on designing the reward and/or goals when the 
outside reward is sparse and hence difficult to learn. Nachum et al.
\cite{nachum2018data} designed a two-level policy, where the
high-level policy produces goals indicating desired changes in the state
observations, and the lower-level policy chooses actions to reach these
specific states. However, this design is limited to applications where we
already know the target goal state, for example, robots for the
locomotion task. Kulkarni et al. \cite{kulkarni2016hierarchical}
proposed to use an intrinsic reward to accelerate learning. In their 
experiments of playing video games, an intrinsic reward is defined as the agent
reaches different objects, and eventually the agent learns the interplay between the 
entities of the system as options. Jaderberg et al. 
\cite{jaderberg2016reinforcement} demonstrated that changes of pixel 
values in the input frames or the learned features in the
hidden layers can be treated as pseudo-rewards and accelerate the learning
when the external reward is sparse. Vezhnevets et al. 
\cite{vezhnevets2017feudal} introduced FeUdal Networks, which is an 
extension of the work by Dayan et al. \cite{dayan1993feudal}. In this
architecture, the manager learns to set sub-goals as directions in the latent 
state space, and the worker learns to fulfill the goals, which can be done using a
mixture of internal and external rewards.

Harb et al. \cite{harb2018waiting} formulate what good options should
be in the bounded rationality framework. Since absolute optimality is
achieved by primitive actions, options are not necessary if we only
focus on achieving optimal final performance. The rational of designing
temporal abstraction is that it can facilitate the learning and achieve a 
better balance of learning speed and final performance.

There are some closely related methods to our work. 
Tsividis et al.\cite{tsividis2017human} investigated human learning in
Atari. They experimentally demonstrated that when humans read
the game's original instruction manual prior to playing, the
first-episode mean score in the game Frostbite increases from 356 to 1848.
They demonstrated that the instruction manual helps humans to form a
model-like representation of the game and enable rapid learning.
Le et al. \cite{le2018hierarchical} imported expert guidance by imitation 
learning to pick the subtasks and their rewards at the coarse (higher) level,
and the agent carried out the reinforcement learning to accomplish
the assigned subtask in the fine (lower) level. Mahjourian et al.
\cite{mahjourian2018hierarchical} played robot
table tennis using a hierarchical control framework. The learning task
is decomposed to a model-free reinforcement learning, a model-based
prediction of ball trajectory, and analytical controllers of the
robot. This strategic decomposition takes advantage of the strengths
of these different approaches. The training and debugging are also
more manageable when only one skill is in focus on at a time. Designing a
real-world robot table tennis is very challenge, and in this work,
after 2 million episodes of cooperative games against itself, the best 
policy can only return the ball about twice, namely the first robot could
not return the ball after the second robot hit it.

\section{Hierarchy from instructions}
The automatic discovery of options that can accelerate the
learning and yet not impose too many constrains to severely jeopardize the
final performance, is one of the hardest problems
in AI research \cite{hassabis2017neuroscience}.  Current main approaches use 
deep reinforcement learning trained end-to-end, in which typically pixel
values are fed in, and end policies are output. This approach is extremely
data thirsty and unnatural to humans. For example, DQN \cite{mnih2015human} 
was trained on 200 million frames for each of the games, which equates
to approximately 924 hours of game time (about 38 days), or almost 500 times
as much experience as the human received. The flexibility of the end-to-end 
approach might seems appealing at first, but the
computation expense is extremely high. This approach is also unnatural to
human learning. For example, it's very unlikely that we let a beginner
drive a car on the road alone to learn how to drive, and 
expect the learner to learn to drive proficiently after many hours of trial-
and-error search, during which the learner has to run over pedestrians many
times to learn the policy of not hitting a pedestrian.

Most of the tasks human faced are accompanied by some form of documentation.
This documentation typically provides an introduction to the basics of the 
environment, objects and targets, rules, and most importantly, a set of
instructions of how to accomplish the task. The perfect example is video game
playing, in which a user manual is expected to be included. In this work, we
decompose the task based on this documentation, and focus on constructing the options
based on the instructions.

In this work, we experimented the idea with the user manual for the Atari video
game of Tennis and Pong. The reason we pick these two game is threefold. First of
all, the entities in this game are just three, namely agent, tennis, and opponent.
The interactions and relationships between these three can be modeled using a linear
model, which will simplify our analysis. Second, they are challenging games to play.
The state-of-the-art result for Tennis achieves 23.6 after 200 million frames, while most other
deep approaches rarely achieve above zero score, namely they cannot beat the embedded AI\cite{hessel2018rainbow}. Third, they have a good analogy 
with playing ball games in the real world. Professional tennis player will learn to
play tennis under the guidance of coaches, and coaches play a key role in sharpening
the learner' skills.
\begin{table}[t]
    \caption{Key instructions from the Atari game Tennis user manual that are
             relevant for the proposed hierarchical approach}
    \label{table-feature}
    \centering
    \begin{tabular}{p{8cm}l}
        \toprule
        Description & Interpretation \\
        \midrule
        \parbox{8cm}{Moving the joystick left moves your player left,
                    moving it right moves him right.} & Action effect.\\
        \midrule
           
        \parbox{8cm}{When you're volleying at the net,
                       your shots don't travel as far as they do
                       when you hit them from the baseline.
                       you'll be able to hit your most sharply-angled
                       shots while playing at the net.} & 
                       \parbox{5cm}{Distance to the
                         net determines hit effect.}\\
        \midrule

        \parbox{8cm}{Aim and "place" the ball depending on how you hit it.
                     The angle of your shot is controlled by where you hit
                     the ball on your racket. If you hit the ball in the
                     center of the racket, your shot will go straight forward.
                     The closer you hit the ball to the edge of the racket, 
                     the sharper the angle will be in the direction your racket
                     is facing. Aim your shots and hit them out of reach of 
                     your opponent.} &
                    \parbox{5cm}{The angle of shots is controlled by where you hit the
                     ball on your racket: hit the ball in the center of racket, 
                     go straight, hit the ball to the edge of racket, the angle will be
                     sharp.}\\
        \midrule

        \parbox{8cm}{Once you understand the game's basics, we suggest you 
                     start off with game 3 or 4 and play in slow motion for a 
                     while. Put the difficulty switches in the b (down) position.
                     You'll soon get the feeling of the court, when to rush the 
                     net, when to lay back and play the baseline.} & 
                    \parbox{5cm}{Understand basics first, practice with easy
                    motions second, and learn higher-level tactics last.}\\
        \midrule

        \parbox{8cm}{Hit a sharply-angled serve off the edge of your racket 
                     to either the right or left side, then move quickly
                     about two-thirds of the way to the net.} &
                    \parbox{5cm}{Switch from hit to wait motion.}\\
        \bottomrule
    \end{tabular}
\end{table}

\subsection{Hierarchy in ATARI video game Tennis}
The original accompanying manual for the Atari video game Tennis
can be found at 
\url{https://atariage.com/manual_html_page.php?SoftwareLabelID=555}.

The key instructions relevant for hierarchy are listed in
Table~\ref{table-feature}. 
These instructions are of tremendous values to build a machine
learning system to play the game Tennis, especially if we want the agent to
learn in a human-like fashion.
\begin{itemize}
    \item Action effect, which indicates we can
          build lower-level skills to move the agent. These can be done using
          supervised learning, reinforcement learning or even analytically. 
    \item The key features in the game planning is that only the hit position 
          matters. First, distance to the net determines the hit effect of
          how long will the ball travel and how sharp-angled will the shot be.
          The shot angle is also controlled by how close the ball is
          hit relative to the center of the racket.
    \item The top-level policy (strategy) of the game becomes deciding when to rush the 
          net, when to play the baseline, and how large the shot angle should be.
    \item The game can be decomposed naturally into two motions, hit and
          wait.
\end{itemize}
Following these instructions, the movement skills are pre-trained first as basic
skills offline, a set of options are designed based on the hit
location (specified by distance to the net and distance of the hit point to
the center of the racket). A hierarchical reinforcement learning
policy is eventually trained to chain these options together to play
the game.

\subsection{Hierarchy in ATARI video game Pong}
The accompanying manual can be found at 
\url{https://boingboing.net/2015/11/07/read-the-manual-to-pong-1976.html}.
Similar to the settings in the game Tennis, in Pong only the hit position
matters. The game task can be naturally decomposed to predict the ball
trajectory and the movement of the paddle. Again, these prediction
models can be pre-trained offline. 

\section{Model and Algorithm}
\subsection{Preliminaries}
Reinforcement learning can be formalized as a Markov Decision
Process (MDP), which can be denoted as a tuple $<S, A, P, R, \gamma>$,
where $S$ is set of states, $A$ is set of actions, $P$ is the state
transition probability, $R$ is the reward function, and $\gamma$ is
the discount. Here we consider episodic tasks in a discrete state space and a
discrete action space. 

Sutton et al. \cite{sutton1999between} 
formulated the hierarchical reinforcement learning by introducing the concept 
of option, which is a generalization of the primitive actions to include 
temporally extended courses of action. An option consist of three components:
intra-option policy $\pi:S\times A \rightarrow[0,1]$, termination condition
$\beta:S^+\rightarrow[0,1]$, and initial set $I\subseteq S$. When
options are temporally extended, the decision process becomes a semi-MDP. 

In this paper we adopt a call-and-return option execution model \cite{bacon2017option}, 
in which the agent chooses an option $o$ according to a
policy over the options $\pi_\Omega(o|s)$, then follows the intra-option policy
$\pi(a|s,o)$ until the termination condition $\beta(s,o)$, and 
these steps are repeated until an episode is terminated. 

\subsection{The State-Option-Reward-State-Option($\lambda$) algorithm}

Compared with Q-learning, which is more popular in Deep Reinforcement
Learning, SARSA is an on-policy method that is stable when combined
with bootstrap and a non-linear model. 
In this work, the SARSA($\lambda$) algorithm from flat reinforcement
learning is generalized to a State-Option-Reward-State-Option
($\lambda$) algorithm for hierarchical reinforcement learning.
Within the hierarchical reinforcement learning framework, option value 
function is defined as 
\begin{equation}
    Q^{\pi_\Omega}(s,o)=E\{r_{t+1}+\gamma r_{t+2}+\gamma^2 r_{t+3}
                           +\dotsm | \epsilon (o\pi_\Omega, s, t)\}
\end{equation}
where $o\pi_\Omega$, the composition of $o$ and $\pi_\Omega$, indicates
the semi-MDP policy that first follows option $o$ until it terminates 
and then chooses the next option according to policy over option $\pi_\Omega$
\cite{sutton1999between}. The semi-MDP version of Q-learning update is:
\begin{equation}
    Q(s, o)\leftarrow Q(s,o)+\alpha \Big[r+\gamma^k \max_{o'\in O_{s'}}
                                 Q(s', o')-Q(s, o)\Big]
\end{equation}
where $k$ denotes the number of time steps elapsed between $s$ and $s'$, and
$r$ denotes the cumulative discounted reward over this time interval.
When the option values are represented via function approximation
parameterized by a weight vector $\mathbf{w}$, it is updated via the update rule
\begin{equation}
    \mathbf{w}_{t+k}\leftarrow \mathbf{w}_t+\alpha \delta_t\mathbf{z}_t
\end{equation}
in which the temporal extended TD error $\delta_t$ of the state-option
value is:
\begin{equation}
    \delta_t=R_{t+1}+\gamma R_{t+2}+ \gamma^2 R_{t+3}+\dotsm
             +\gamma ^{k-1}R_{t+k}
             +\gamma^k Q(s_{t+k}, o_{t+k}, \mathbf{w}_t)
             -Q(s_t, o_t, \mathbf{w}_t)
\end{equation}
and the state-option value form of the eligibility trace is:
\begin{eqnarray}
    \mathbf{z}_1 &=& \mathbf{0} \nonumber \\
    \mathbf{z}_{t+k} &=& (\gamma \lambda)^k
                      \mathbf{z}_t+\bigtriangledown Q(s_t, o_t, \mathbf{w}_t)
\end{eqnarray}
in which $\lambda\in[0,1]$ is trace decay parameter
\cite{sutton2018reinforcement}.

\begin{algorithm}
    \caption{SORSO($\lambda$) algorithm}
    \label{alg:sorso}
    \begin{algorithmic}
        \STATE $\mathbf{w}\leftarrow \mathbf{0}$ 
        \STATE $\epsilon =1$ 
        \FOR{$ep=1$ to $N^{Episodes}$}
        \STATE $\mathbf{z}\leftarrow \mathbf{0}$
        \STATE $R=0$ \COMMENT{ R is the discounted cumulative reward}
        \STATE $k=0$
        \STATE Reset game and get start state $s$
        \STATE Choose option $o$ with $\epsilon$-greedy
        \STATE $a\leftarrow \pi(s,o)$
        \WHILE{$s$ is \NOT terminal state}
        \STATE Execute $a$ and obtain next state $s'$ and reward $r$
               from environment
        \STATE $k\leftarrow k+1$
        \STATE $R\leftarrow \gamma R+r$
        \IF{ $s'$ is terminal}
        \STATE $\delta=R-Q(s, o, \mathbf{w})$
        \STATE $\mathbf{z} = (\gamma \lambda)^k \mathbf{z}
                             +\bigtriangledown Q(s, o, \mathbf{w})$
        \STATE $\mathbf{w}\leftarrow \mathbf{w}+\alpha \delta\mathbf{z}$

        \ELSIF{$o$ terminates at $s'$}
        \STATE Choose option $o'$ with $\epsilon$-greedy
        \STATE $a'\leftarrow \pi(s',o')$
        \STATE $\delta=R+\gamma^k Q(s', o', \mathbf{w})-Q(s, o, \mathbf{w})$
        \STATE $\mathbf{z} = (\gamma \lambda)^k \mathbf{z}
                             +\bigtriangledown Q(s, o, \mathbf{w})$
        \STATE $\mathbf{w}\leftarrow \mathbf{w}+\alpha \delta\mathbf{z}$
        \STATE $s \leftarrow s'$
        \STATE $o \leftarrow o'$
        \STATE $a \leftarrow a'$
        \STATE $k=0$

        \ELSE
        \STATE $a \leftarrow \pi(s',o)$
        \STATE $s \leftarrow s'$
        \ENDIF
        \STATE Anneal $\epsilon$
        \ENDWHILE
        \ENDFOR
    \end{algorithmic}
\end{algorithm}
In Algorithm ~\ref{alg:sorso}, the sub-policy $\pi(s, o)$ is a pre-trained
skill, which will be discussed more in the Experiments Section. 

\section{Experiments}
\subsection{Atari game Tennis}
\begin{figure}
    \centering
    \includegraphics[width=0.4\textwidth]{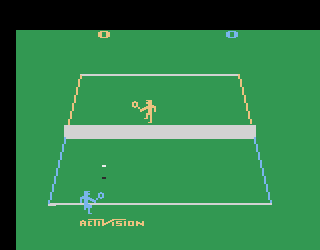}
    \caption{A frame from the game Tennis.}
\label{fig:screen-tennis}
\end{figure}

\paragraph{Setup}
The Atari game Tennis (see Figure ~\ref{fig:screen-tennis}) 
is played with both hierarchical and flat policy reinforcement learning.
Using the instructions from the handbook, a hierarchical policy and set of 
options are designed. When the ball starts flying towards the agent, a
hit mode is activated. In the hit mode, the option is designed as 
$[\Delta x, \Delta y]$, in which $\Delta x$ is the distance between the hit
position and the base of the net, and $\Delta y$ is the difference between the
hit position and the center of the racket. These values are discretized to $\Delta x \in \{0, \frac{L}{4}, 
\frac{2L}{4}, \frac{3L}{4}, \frac{4L}{4}\}$, in which $L$ is the
half length of the tennis court, and $\Delta y \in \{-15, 10, -5, 0, 5,
10, 15\}$. The values of $\Delta x$ are chosen such that they cover both ends of the court and intermediate locations between the two ends.
The values of $\Delta y$ are chosen heuristically and cover
the spectrum from dangerously close to the racket edge (which leads to a highly skewed trajectory)
to center (which leads to a straight shot). 

The wait mode is activated when the 
agent hits back the ball, and the option moves the
agent to two thirds of the way to the net, and the middle between the
sidelines. In the wait mode, if the agent has not arrived at the
designated location but detects the tennis has been hit back, it will
immediately switch to hit mode.
To carry out options, a key observation is that the trajectory of the ball
needs to be predicted accurately. In this game design, when the ball is flying, its velocity on the horizontal surface is fixed, so the
projection of the ball trajectory on the court is a straight line and
we can analytically predict the projection with zero error. Another 
component is the low-level skill of moving to the designated location.
In this game, we always try to move to the target location as soon as
possible. The action effect can be pre-trained offline, and in this 
game design, the movement of one step of action is fixed, so the
subpolicy $\pi(s, o)$ can be derived analytically here.

\paragraph{Training}
In this experiment, we used the Gym toolkit \cite{brockman2016openai} as 
the simulation bed. The positions of objects (namely agent, ball, and 
opponent) and the ball velocity comprise the system state, and a set of 
Fourier features \cite{konidaris2011value} are calculated based on this 
state. The learning rate is set to be $5.0\times 10^{-5}$, discount ratio
$\gamma=0.99$, and eligibility trace decay factor $\lambda=0.99$. The
exploration parameter $\epsilon$ is decreased exponentially. To make a
comparison with the flat reinforcement learning, a flat policy is
also trained here with the SARSA($\lambda$)\cite{sutton2018reinforcement}
algorithm.

\paragraph{Results}
\begin{figure}
    \centering
    \includegraphics[width=0.45\textwidth]{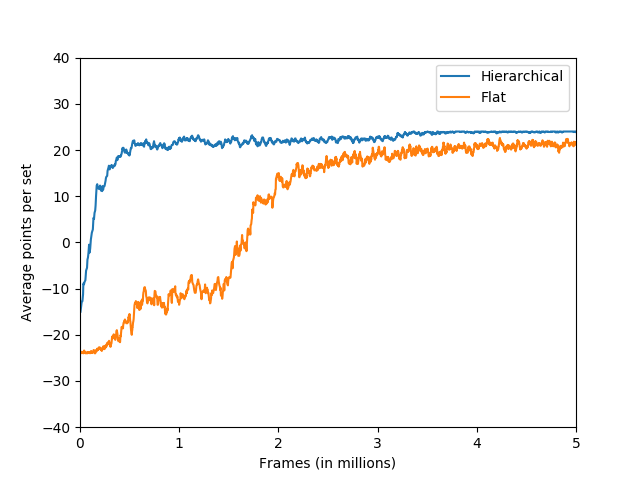}
    \caption{Learning curve for the game Tennis. Each curve is smoothed with a 10-set
    moving average.}
\label{fig:learning-curve}
\end{figure}
The learning curves are plotted in Figure ~\ref{fig:learning-curve}.
In hierarchical reinforcement learning, after about 4 million frames, the
agent achieved the upper bound of its performance, 24, namely wins all the
points in a single set. The agent achieved the human level performance of $-8.9$
after about 0.1 million trained frames, while a human player needs about
two hours of practice, which amounts to around 0.4 million frames on a
60 Hz emulator\cite{mnih2015human}, to achieve this level of performance. 
The agent learned to play the game with human-level efficiency and quickly
outperforms humans. The hierarchical policy learns an order of magnitude faster than the flat policy. 
The agent also outperforms an end-to-end deep reinforcement
learning system by a large margin with respect to learning efficiency
(5 million vs. 200 million frames trained\cite{hessel2018rainbow}).

\subsection{Atari game Pong}
\paragraph{Setup}
Instructed by the manual, the hit mode and wait mode are designed just as in Tennis. 
In the hit mode, the option is designed as $\Delta y$, and $\Delta y$
is the difference between the hit position and the center of the agent 
paddle, $-8 \leq \Delta y \leq 8$. For reference, the height of paddle is 
8, and height of the ball is 2, so values of $\Delta y$ cover all the possible
hit positions. When the ball is hit to the opponent, the wait mode is 
activated and action of no-op is imposed until the ball gets hit back by the
opponent. Since the agent can move swiftly, the paddle does not need to
be moved to the center in wait mode.

To implement options, a model for the movement of the ball and movement of the
paddle has to be built. In this game design, when the ball is flying, 
the velocity is fixed, so we can predict it precisely. The low-level skill
of moving to the designated location is built on the action effects on the 
paddle, which is pre-trained offline.
\begin{figure}
    \centering
    \includegraphics[width=0.4\textwidth]{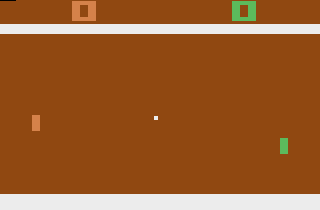}
    \caption{A frame from the game Pong.}
    \label{fig:screen-pong}
\end{figure}
\paragraph{Training}
The same setting as in playing Tennis is adopted here. A predictive
model for the action effect is also trained using supervised learning. The 
past 4 frames of agent paddle and the past 3 actions are recorded to
predict the action effect. To collect the samples, a random policy is
\begin{figure}
    \centering
    \includegraphics[width=0.45\textwidth]{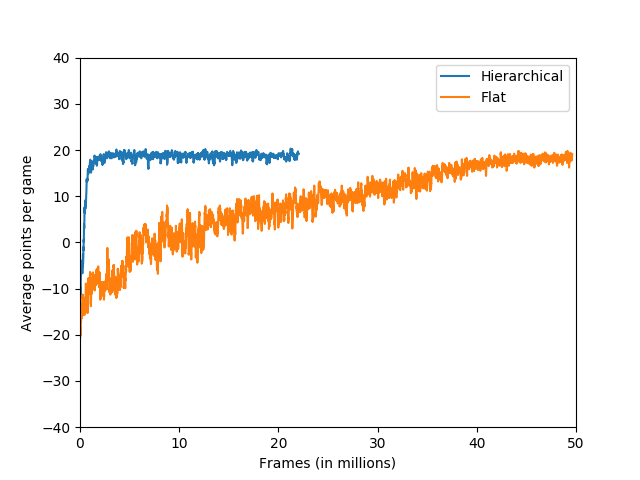}
    \caption{Learning curve for the game Pong. Each curve is smoothed with a 
            10-game moving average.}
    \label{fig:learning-curve-pong}
\end{figure}
implemented and 1 million unique samples are collected. After
training, a training error of 1.06 is achieved for the action ``no-op'', 1.47 
for the action ``up'', and 1.52 for the action ``down''. It's noteworthy that the
predictive model is not perfect. A careful inspection on the
samples indicates that the agent paddle moves with significant randomness.
Although the predictive model cannot be precise, experiments show
that it is adequately reliable for the top-level policy.
In this study, the prediction model for the agent movement effect and model for the
ball trajectory prediction are constructed explicitly and in isolated manner.
In end-to-end reinforcement learning, all these sub-skills are learned
together implicitly. With the diverse entities involved and entangled, 
it will be much more difficult to learn them all at once.

\paragraph{Results}
The learning curves are plotted in Figure
~\ref{fig:learning-curve-pong}.
In hierarchical reinforcement learning, after about 0.6 million frames, the 
agent achieved human level performance of 9.3\cite{mnih2015human}, and a
human needs about 0.4 million frames of practicing to achieve this score.
The hierarchical policy learns an order of magnitude faster than the flat policy, and got a
final performance around 19, while the flat policy got a final performance 
about 18.3. The hierarchical policy outperforms an end-to-end deep reinforcement learning system
in efficiency by an order of magnitude, while the flat policy learns with about the same speed
as deep learning.

\section{Conclusion}
In this work we propose that machines can learn from instructions
provided by handbooks and/or expert human, and these instructions can
provide hierarchical decomposition of the task into components, which
can be tackled by different learning systems to take advantage of their
strengths. Instruction manuals also provide an abstract illustration of the
key features of the environment, such as guidelines of the 
spatial-temporal structure of the planing procedure, to accelerate the
exploration of the environment. Low-level skills can also be advised by
instructions to build supporting models for rapid learning.
The experiments indicate that using instructions from handbook, machines
can learn as fast as human and reach human-level performance with the
same amount of training data, and outperform humans after more training. The
building of the supporting models and the appropriateness of the option
construction are the two key factors in the learning ability of the
hierarchical agent.

\small

\bibliographystyle{plain}
\bibliography{references}

\end{document}